# BMDetect: A Deep Learning Framework for Comprehensive Biomedical Misconduct Detection

Yize Zhou[1], Jie Zhang[2], Meijie Wang[3], Lun Yu[3*]

yizezhou20001203@163.com

*Corresponding author: lunyu@metanovas.com

1. School of Mathematics and Statistics, Wuhan University, Wuhan, China

2. School of Information and Software Engineering, University of Electronic Science and Technology of China, Chengdu, China

3. Metanovas Biotech Inc., San Francisco, California, USA

**Abstract**

Academic misconduct detection in biomedical research remains challenging due to algorithmic narrowness in existing methods and fragmented analytical pipelines. We present BMDetect, a deep learning framework that integrates journal metadata (SJR, institutional data), semantic embeddings (PubMedBERT), and GPT-4o-mined textual attributes (methodological statistics, data anomalies) for holistic manuscript evaluation. Key innovations include: (1) fusion of domain-specific features to reduce detection bias; (2) quantitative evaluation of feature importance, identifying journal authority metrics (e.g., SJR-index) and textual anomalies (e.g., statistical outliers) as dominant predictors; and (3) the BioMCD dataset, a large-scale benchmark with 13,160 retracted articles and 53,411 controls. BMDetect achieves 74.33% AUC, outperforming single-modality baselines by 8.6%, and demonstrates transferability across biomedical subfields. This work advances scalable, interpretable tools for safeguarding research integrity.

## Introduction

In recent years, biomedical research has been pivotal in driving scientific breakthroughs and advancing medical practices. High-quality biomedical publications are essential for clinical decision-making, policy formulation, and drug development [1]. However, academic misconduct, often driven by low risks and potentially high rewards, has been increasingly prevalent and is a focus of public and governmental concern. According to the U.S. Office of Science and Technology Policy [2], misconduct involves actions that significantly deviate from accepted norms, including fabrication, falsification, and plagiarism, while excluding honest errors or differences in opinion. Such unethical practices are particularly rampant in biomedical research. Statistics indicate that out of approximately one million new articles listed annually in PubMed, up to 90% lack reproducibility [3,4]. A considerable portion of this irreproducibility stems from deliberate fabrication or falsification of results [5]. The consequences include wasted research time and financial resources [6], diminished public trust in scientific endeavors [7], and potential threats to public health and safety [8].

Academic misconduct in the biomedical field manifests in various forms, each posing significant threats to scientific integrity and progress. One prominent type is data fabrication, where researchers invent experimental data or results to support specific conclusions, thereby undermining scientific credibility and reproducibility [9]. Another is data falsification, which involves selecting or omitting data to favor a hypothesis, such as excluding unexpected results or altering statistical analyses. This practice can mislead subsequent research and clinical applications [10]. A third is plagiarism, which includes copying text, figures, or data without proper attribution, or self-plagiarism through unauthorized republication, violating ethical and legal standards in biomedical research [11]. Given the complexity and prevalence of these issues, identifying misconduct is imperative. Although commonly used, manual manuscript analysis is increasingly impractical due to the sheer volume of biomedical publications [12]. This underscores the urgent need for automated tools to detect and evaluate academic misconduct efficiently and accurately.

In recent years, the rapid development of artificial intelligence technologies has sparked significant interest in applying machine learning (ML) and deep learning (DL) to detect academic misconduct. These methods have been utilized to identify various forms of unethical behavior, such as text plagiarism, image manipulation, and data falsification, demonstrating notable advantages and potential. At the textual level, academic misconduct detection has been a core focus. For example, Mohamed A. El-Rashidy et al. developed a plagiarism detection system based on feature selection and support vector machines (SVM), emphasizing feature extraction to improve detection performance [13]. The system demonstrated exceptional accuracy and efficiency, supporting the automated detection of textual similarity and misconduct. Similarly, Asif Ekbal and colleagues employed multi-level natural language feature processing and similarity analysis, proposing an external plagiarism detection technique that effectively handled paraphrasing, achieving high accuracy and practical applicability [14]. With the rise of deep learning, natural language processing (NLP) has become integral to academic text analysis. Nguyen et al. proposed a two-stage plagiarism detection method using LSTM networks and feature extraction: in the paragraph stage, article similarity and feature extraction located plagiarized articles, while in the word stage, specific content was identified through word and sentence similarity. This method achieved notable success in textual plagiarism detection [15]. Researchers have also developed intelligent systems combining convolutional and recurrent neural networks to analyze lexical, syntactic, and semantic features in databases, achieving superior detection accuracy on datasets like PAN 2013 and PAN 2014 [16].

In image manipulation, issues like image reuse and forgery significantly compromise the credibility of research outcomes. While tools like Forensically [17] and Imagetwin [18] assist reviewers in pixel-level forensics, they often rely on human interpretation and experience, reducing reliability. Machine learning and deep learning approaches have advanced the detection of image forgery. Thakur et al. introduced a method using discrete wavelet transform (DWT) to decompose images into sub-images, extracting coefficients with robust feature extraction for forgery classification via SVMs [19]. Emad Ul Haq Qazi et al. proposed a spliced image forgery detection method utilizing ResNet50v2 and YOLO CNN weights. The model achieved high accuracy through transfer learning, trained and tested on CASIA_v1 and CASIA_v2 datasets, showcasing a significant improvement over existing methods in detecting spliced forgeries [20].

Furthermore, data falsification poses substantial challenges to the integrity of academic research, with traditional anomaly detection methods, such as statistical analysis, facing limitations in handling complex patterns. Malhotra Pankaj et al. proposed an LSTM-based anomaly detection model capable of identifying potential falsification by analyzing temporal patterns in data [21]. The introduction of transformer architectures has opened new possibilities for multidimensional data analysis. For instance, Wang et al. developed an unsupervised anomaly detection model based on variational transformers. This model leveraged the self-attention mechanism and enhanced position encoding with upsampling algorithms to capture inter-sequence relationships and multi-scale temporal information, integrating residual variational autoencoders to achieve efficient anomaly detection [22].

A review of prior research highlights a few significant gaps in addressing academic misconduct in the biomedical domain. The first is the limited research focus. Few studies have developed comprehensive tools specifically tailored to the unique challenges of the biomedical field. And the second is methodological narrowness. Existing methods often focus on isolated issues, such as plagiarism, data fabrication, or methodological inconsistencies, without providing a holistic evaluation of entire manuscripts. Furthermore, the detection process frequently involves analyzing multiple elements across several steps, which increases complexity and limits scalability. Addressing these gaps is essential for developing effective and efficient tools to combat misconduct in biomedical research.

This study proposes a novel approach leveraging the BERT framework to address these challenges. A curated dataset was constructed by screening Retraction Watch [23] and PubMed databases, labeling retracted articles with misconduct (fabrication, falsification, plagiarism) as positive samples, and selecting negative samples at a 1:3 ratio. Features were derived by combining textual attributes, such as journal-specific properties, with deep semantic patterns extracted from the text using GPT-4o. This integration captures explicit information and nuanced latent features within the data, enabling a more robust detection model. The final detection model integrates PubMedBERT-based pre-trained title features, GPT-4o-derived latent textual attributes, and journal features. we propose the Biomedical Misconduct Detector (BMDetect), which introduces three key innovations:

1. Heterogeneous Feature Integration: Synthesizes structured metadata (journal SJR, institutional networks), deep semantic features (PubMedBERT embeddings), and methodological statistics (data dispersion, method citation density) to overcome single-feature detection biases;
2. Feature Importance Evaluation: Quantifies the contributions of journal authority metrics (e.g., SJR-index correlations) and textual anomalies (e.g., statistical outliers in results sections) through rigorous importance ranking, identifying dominant predictors of misconduct.
3. Curated Dataset Construction: Develop the Biomedical Misconduct Detection Dataset (BioMCD), which includes 13160 retracted articles and 53411 non-retracted control articles. The construction method ensures a high correlation between positive and negative samples, thus forming a high-quality dataset.

# Methods

## Dataset construction

Constructing a dataset is a pivotal step in deep learning to ensure the efficacy of model training. To develop a model capable of accurately predicting the likelihood of paper retraction, we utilized the API provided by Crossref Labs to obtain data from the Retraction Watch database [24] of retracted papers. This database meticulously documents retracted articles along with their reasons for retraction. Our study specifically focuses on the category of "Health Sciences."

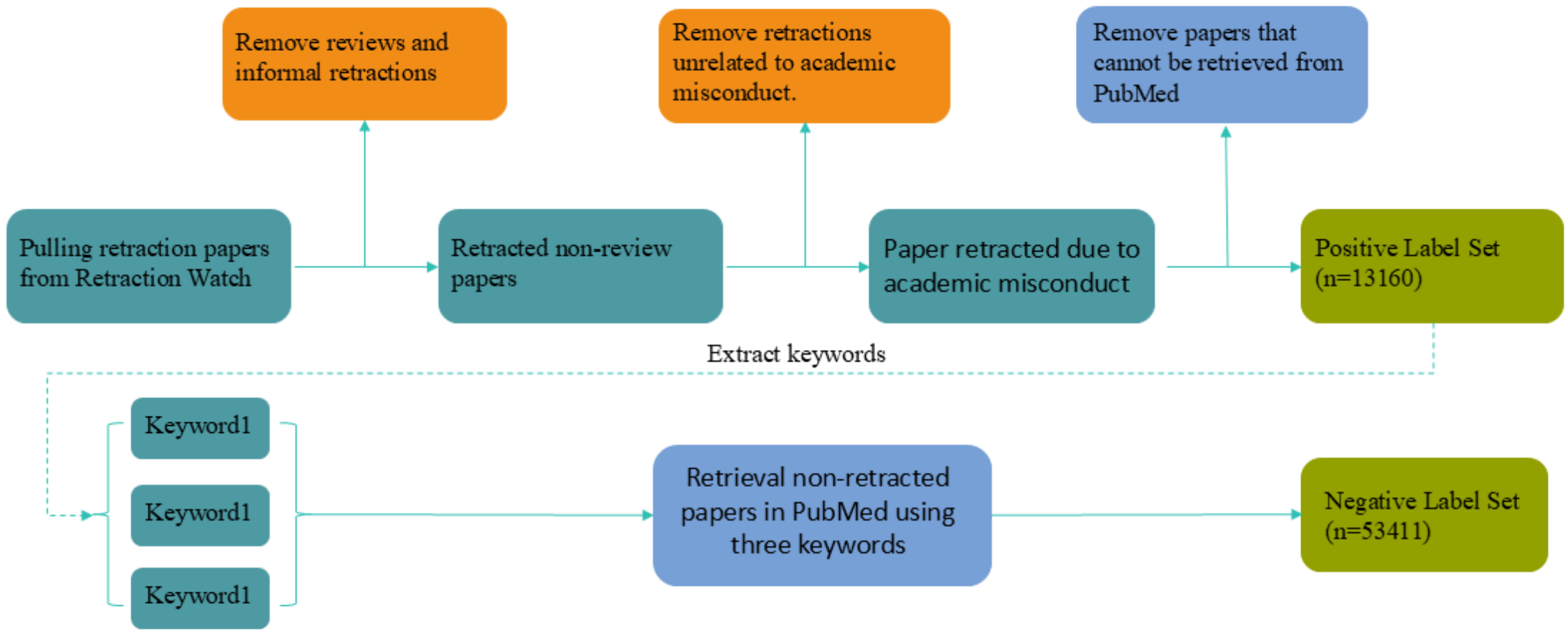

**Fig 2 Data processing flowchart**

The initial dataset comprised 53,716 records. During the preprocessing phase, we refined this dataset by excluding entries with retraction statuses marked as "Correction" and "Expression of Concern", since these articles, while problematic to some extent, have not undergone formal retraction and could thus introduce noise into the model training process. Our study aims to identify papers at risk of retraction due to academic misconduct. In academia, such misconduct primarily manifests as fabrication, falsification, and deception. Therefore, we select only those articles from the original dataset with clear retraction reasons involving fabrication, falsification, and deception related to academic misconduct to construct our experimental dataset. This approach ensures the focus and accuracy of the dataset, facilitating a more in-depth analysis of the impact and characteristics of these unethical behaviors. Furthermore, review articles were also removed from the dataset to ensure that all samples consisted solely of original research papers. This exclusion criterion was applied to enhance the relevance and accuracy of model predictions.

Consequently, we arrived at a refined dataset focused on the "Health Sciences" domain, containing 26,344 formally retracted research articles. This curated dataset provides a solid foundation for subsequent feature engineering and model training.

We obtained the PubMed ID for each article from our existing retraction dataset to construct a more comprehensive dataset of retracted papers. Leveraging these IDs, we queried the PubMed Central (PMC) database [25] via the PubMed API to retrieve the full text of each retracted paper. PubMed Central (PMC) is an archive maintained by the National Library of Medicine (NLM) at the National Institutes of Health (NIH), which provides free access to full-text articles in the fields of biomedicine and life sciences.

After removing the records that did not retrieve the full text of the papers, we constructed a positive label set. Each record in this set contains the full text of retracted papers, ultimately comprising 13,160 papers.

Upon completing the construction of the positive label set, we randomly selected three keywords from the keyword section of each retracted paper's full text. For papers that did not provide keywords in their full text, we used openchat-3.5 to summarize keywords reflecting the research content based on their titles. Using the keywords obtained as described above, we searched for full texts of non-retracted papers that contain these keywords and were published within one year of the corresponding retracted papers, using the PubMed API from the PubMed Central (PMC) database. Our objective is to match three non-retracted control papers for each retracted paper in the positive label set. This is because, in many scientific research papers, particularly in the life and medical sciences, statistical analyses have shown that approximately 75% [26,27] of articles exhibiting academic misconduct are identified during the publication process. In this study, the papers exhibiting academic misconduct are considered positive samples, and based on this proportion, we have designed the dataset of positive and negative samples for this research.

To ensure the quality and completeness of the dataset, we removed all records that could not be obtained in full text during the retrieval process. Ultimately, we constructed a negative label set comprising 53,411 non-retracted papers. This process not only ensured a high degree of relevance between the positive and negative samples but also minimized biases due to temporal differences as much as possible, thereby providing high-quality data support for subsequent analyses.

To optimize model training, we further extract information from the collection of papers labeled as positive and negative classes. The full text of these papers is provided in XML format. From this, we extract each paper's title, publication year, and the journal in which it was published. By enriching the feature dimensions of the dataset, this process not only enhances the depth of the dataset content but also effectively supports the accuracy and efficiency of model training.

## Feature Construction

In detecting academic misconduct in biomedical literature, feature selection and construction are critical to the model's effectiveness. We created and trained a deep-learning model from scratch based on these selected features. For this task, we integrated multidimensional features derived from pre-trained models, large language models (LLMs) that mine potential textual attributes, and journal- and paper-related information to construct the final detection model. The details of feature construction are as follows:

1. Title Features

The title is a crucial component of a paper, reflecting its core research content and topic. We used PubMedBERT [28], a pre-trained model, to extract embedded features from titles. PubMedBERT is optimized for biomedical text and trained on millions of biomedical documents, including titles, abstracts, and full texts. It effectively captures the semantic and academic properties embedded in titles. By processing the titles of the collected literature (details on the extraction method are described in the model architecture section), we obtained high-quality semantic features that provide critical support for detecting academic misconduct.

2. LLM-Based Features

We leveraged GPT-4o [29] to analyze the full-text data contained in the collected XML documents. Considering that academic misconduct often manifests as fabrication of facts in the "Methods" and "Results" sections, or falsification of experimental data, we focused on these sections. Using GPT-4o, we mined and extracted key features related to the "Methods" and "Results" sections, including:

(1) Method Features:

- Method_Num. The number of methods mentioned in the "Methods" section.
- Method_Cite. The number of citations in the "Methods" section.
- Method_Statistical. The number of statistical terms mentioned in the "Methods" section.

(2) Result Features:

- Result_Fig. The number of figures or tables in the "Results" section.
- Result_Num. The number of data entries in the "Results" section.
- Result_Statistical. The number of statistical results reported in the "Results" section.

3. Full Text Features

The foundational information about journals and papers serves as an essential dimension for evaluating the quality and credibility of literature. Based on the XML data, we retrieved or constructed the following features:

(1) Journal-related Features:

- Scientific Journal Ranking (SJR) [30]. The SJR impact factor of the journal. This indicator reflects the quantity and quality of citations, measuring the journal's scientific influence and academic reputation. SJR (SCImago Journal Rank) is calculated by weighting the number of citations a journal receives based on the prestige of the citing sources, making it a critical metric for evaluating journal impact.
- The H-Index [31]. The h-index is a composite metric that indicates the number of articles (h) published by a journal that have been cited at least h times, simultaneously assessing productivity and citation impact. A higher h-index suggests greater academic output and influence.
- Quartile_Journal [32]. The quartile ranking of the journal (e.g., Q1, Q2). A journal's influence and SJR ranking within its research field determine quartiles by dividing journals into four groups. Q1 represents the top 25% of journals, recognized as highly influential in academia.
- Country_Journal. The country where the journal is based. This feature reflects the geographical distribution of journals, providing insights into their internationalization and the academic influence of the publishing country. Variations in academic standards, research focus, and resource allocation may exist across different countries.
- Area_Journal. The research area of the journal. This feature indicates the journal's disciplinary scope or research focus, such as pharmacology, toxicology, etc. Classification of research areas helps the model assess the thematic relevance of the literature based on the journal's academic background.
- Year. The year the paper was published. The publication date is an important parameter for evaluating the timeliness of research and citation patterns. In rapidly evolving fields, newer literature often contains more cutting-edge findings.

(2) Paper-related Features:

- Is_Clinical. Whether the paper is a clinical study. This feature identifies whether the paper is a clinical study, commonly associated with research in medicine, pharmacology, and related fields. Clinical studies evaluate the safety and efficacy of drugs, medical devices, or interventions in human subjects. Such research is crucial for advancing medical practice and improving patient outcomes.
- Human/Animal/Molecular_Cellular [33]. The type of research subjects (e.g., human, animal, or molecular/cellular).

(3) Institution and Author Information:

- Affiliation. Detailed information about the authors' affiliations. This feature provides comprehensive details about the institutions the authors are affiliated with, including the institution name, department, or laboratory. These affiliations reflect the research environment and resources available to the authors and can influence the scope and quality of the research conducted.
- Aff_Countrys. The countries of the institutions. This feature identifies the countries where the authors' affiliated institutions are located. The scientific capabilities and research focus of different countries can significantly impact the academic influence of a paper. Developed countries often have more resources and higher-quality research outputs.
- Aff_Areas. The research fields of the institutions. This feature describes the research areas

of the authors' affiliated institutions, such as life sciences, engineering, or social sciences. These areas provide insight into the academic context and research focus of the paper, reflecting the institution's strengths or specialization.

- Aff_Natures. The nature of the institutions (e.g., research institute, business, university).

By constructing the above features, we integrated semantic features, textual attributes, journal characteristics, and institutional information to support the development of the academic misconduct detection model. The diversity and rationality of the constructed features significantly enhance the model's ability to identify instances of academic misconduct.

## BMDetect model

BMDetect model consists of three modules: an intra-sample text feature encoder that converts the title string into an embedding vector, an attention encoder that transforms the intra-sample text feature vectors of a single sample into an inter-sample weighted representation, and a journal feature encoder that extracts information generated by large models and article journal features.

The text encoder module focuses on the representation of textual features within each sample. For each sample $i$, we use the PubMedBERT pre-trained model to tokenize the title text into a sequence $T_i$. This sequence is then fed into PubMedBERT for encoding, where the attention mechanism allows the embedding vectors corresponding to each word to aggregate information from all words in the text. In the final hidden layer, the embedding associated with the retained [CLS] token is used for downstream classification tasks [34]. We define $b_i$ as the intra-sample vector representation of $T_i$, which is mathematically expressed as follows:

$$b_i = \text{PubMedBERT}\left(T_i\right)$$

We define $d$ as the embedding size, where $b_i \in \sim^{d}$. It is important to note that $b_i$ is associated with the [CLS] token and contains information about the entire text.

The second module in our model is an attention encoder, which focuses on the representation of inter-sample text features. By applying the attention mechanism, we re-encode the title vectors of each article into weighted title vectors, allowing the model to learn subtle differences between the title representations of multiple articles. Let the batch size be defined as $s$, representing the number of samples processed in each training epoch. Let $b = \left[b_1, b_2, \cdots, b_s\right]^T$ denote the intra-sample title vector representation for the entire batch. We represent the query (Q), key (K), and value (V) matrices as $Q, K, V \in \sim^{d\times d}$, where $d$ is the embedding dimension. By multiplying with the training matrices $W_Q$, $W_K$, and $W_V$, we obtain: $Q = bW_Q$, $K = bW_K$, $V = bW_V$. We define $\alpha_i$ as the computed weight vector for sample $i$, which is represented by the text vectors of other samples in the same batch. The formula for the title text feature $B_i$ is as follows:

$$\alpha_i = \text{Softmax}\left(q_i K^T\right)$$

$$B_i = \alpha_i \cdot V$$

In this way, the model can capture the relationships between samples, enhancing its ability to understand and represent text features. In an ideal scenario, when applying the attention mechanism to compute the inter-sample vector weights, each sample in the dataset should be considered. However, due to limited computational resources, we construct a sample set from each batch. As a result, all text vectors within the same batch are weighted to obtain the inter-sample representation for each vector.

The third module is the journal feature encoder, which processes information generated by large language model and journal-related features. Categorical features are converted into dummy variables, while continuous features are normalized. All categorical and continuous features are concatenated to form the complete journal feature vector $j_i$ for sample $i$. These features are then passed through a feedforward layer for a linear combination of the different journal features. Finally, the output features are input into the ReLU activation function, commonly used in deep neural networks.

For sample $i$, we obtain its inter-sample title text features $B_i$ and enhanced journal features $J_i$. By concatenating these, we obtain the overall feature vector $X_i$ for final classification, where $X_i = \text{concat}\left(B_i, J_i\right)$. For a batch with $s$ samples, $X = \left[X_1, X_2, \cdots, X_s\right]$, and the predicted labels are represented as $Y$. In the fully connected layer, we have:

$$Y = \text{Softmax}\left(WX + c\right)$$

where $W \in {\sim}^{\tilde{d}} \times {\sim}^{\tilde{d}}$, and $\sim$ representative twice the embedding size plus the journal feature size.

# Results

## Models and Metrics

We conducted a series of experiments to evaluate the classification performance of the proposed model. For comparison purposes, we considered several competitors, including XGBoost, Convolutional Neural Networks (CNNs), Recurrent Neural Networks (RNNs), other attention mechanisms, and pre-trained language models. The text and journal feature inputs for all models were similar.

- XGBoost [35]: In this model, the embedded title features, the features extracted from the large-scale model, and the journal-related features are concatenated and fed into an XGBoost classifier with 100 decision trees for training and prediction. XGBoost, an optimized gradient boosting framework, constructs decision trees iteratively, where each tree corrects the residual errors of the previous ones. This process enhances the model's ability to capture complex feature interactions and improves classification performance. Through careful hyperparameter tuning, XGBoost effectively balances model complexity and generalization, ensuring robust predictions.

- TabNet [36]: In this model, the embedded title features, features extracted from large models, and journal features are concatenated and fed into a TabNet classifier for training and prediction. TabNet utilizes an attention mechanism to select features and models them through a decision tree structure. The hyperparameters are set as follows: n_d=8 (feature dimension in the decision block), n_a=8 (feature dimension in the attention block), n_steps=3 (number of decision steps), gamma=1.3 (independence control parameter), and lambda_sparse=1e-4 (sparsity regularization term). The model is evaluated using 5-fold cross-validation, tracking accuracy, F1 score, AUC, precision, and recall metrics.
- BiLSTM [37]: This is a type of Recurrent Neural Network (RNN) that excels in text classification tasks, such as sentiment analysis and question classification. The text encoder consists of a two-layer BiLSTM, and its output hidden states are concatenated with the journal feature representations, which are processed using simple fully connected layers. The word embedding dimension is set to 128, and the BiLSTM hidden state dimension is 256. A fully connected layer followed by a SoftMax function is applied to produce the final prediction.
- BiLSTM+Attention [38]: This model builds upon the standard BiLSTM architecture by integrating an additional attention layer at the output of the text encoder. This attention mechanism enables the model to dynamically assign weights to different parts of the encoded text sequence, thereby emphasizing the most informative features and capturing richer contextual information. The enhanced feature representation ultimately leads to improved performance in text classification tasks.
- TextCNN [39]: This Convolutional Neural Network (CNN) is used for sentence classification tasks, with a kernel size of 32 to extract sentence-level features. For the journal feature representations, we employed the BiLSTM method and fed the concatenated representations into the classifier.

To ensure the stability of our results, we employed a 5-fold cross-validation approach instead of a simple train-test split. The dataset was randomly partitioned into five subsets, with each subset serving as the test set once while the remaining four were used for training. This process was repeated five times, and the final performance metrics were averaged across all folds. We evaluated model performance based on true positives (TPs), representing correctly predicted positive samples; true negatives (TNs), indicating correctly predicted negative samples; false positives (FPs), referring to negative samples incorrectly classified as positive; and false negatives (FNs), which are positive samples misclassified as negative. Using these values, we computed key evaluation metrics, including Precision = TP / (TP + FP), Recall = TP / (TP + FN), Specificity = TN / (FP + TN), Accuracy = (TP + TN) / (TP + FP + FN + TN), F1-score = 2 × [(Precision × Recall) / (Precision + Recall)], and Area Under the Receiver Operating Characteristic Curve (AUC) = $\int_0^1 Recall d\left(Precision\right)$. The 5-fold cross-validation approach reduces variance in performance estimates and ensures that the evaluation results are more reliable and generalizable.

## Experimental results

Table 1 Experimental results of different models when evaluated by Precision, Recall, Specificity, Accuracy, F1 Score, and area under the receiver operating characteristic curve (AUC)

| Model | Precision | Recall | Specificity | Accuracy | F1-score | AUC |
|---|---|---|---|---|---|---|
| XGBoost | **0.6453 ±** | 0.3451 ± | **0.9060 ±** | **0.7200 ±** | 0.4494 ± | 0.7373 ± |
| | **0.0112** | 0.0217 | **0.0042** | **0.0058** | 0.0204 | 0.0079 |
| TabNet | 0.6082 ± 0.0322 | 0.3142 ± 0.1492 | 0.9054 ± 0.0435 | 0.7093 ± 0.0210 | 0.3902 ± 0.1749 | 0.6859 ± 0.0830 |
| BiLSTM | 0.4586 ± 0.0133 | 0.6890 ± 0.0217 | 0.5959 ± 0.0232 | 0.6268 ± 0.0141 | 0.5505 ± 0.0127 | 0.6967 ± 0.0155 |
| BiLSTM+Attention | 0.4780 ± 0.0108 | 0.6282 ± 0.0642 | 0.6581 ± 0.0462 | 0.6482 ± 0.0107 | 0.5409 ± 0.0208 | 0.6984 ± 0.0082 |
| TextCNN | 0.4476 ± 0.0045 | 0.6673 ± 0.0289 | 0.5915 ± 0.0164 | 0.6166 ± 0.0049 | 0.5356 ± 0.0108 | 0.6791 ± 0.0099 |
| Our Model | 0.4824 ± 0.0121 | **0.7403 ± 0.0407** | 0.6045 ± 0.0404 | 0.6495 ± 0.0138 | **0.5833 ± 0.0048** | **0.7433 ± 0.0057** |

The experimental results of various models evaluated through 5-fold cross-validation are presented in Table 1, including precision, recall, specificity, accuracy, F1 score, and AUC index statistical information (mean ± standard deviation, i.e. Mean ± SD).. Our proposed model demonstrates superior performance in critical metrics, particularly in detecting positive cases, making it highly suitable for academic misconduct identification tasks. The key advantages of our model are as follows:

- **Highest Recall (0.7403 ± 0.0407)**: Our model significantly outperforms all baseline models in Recall, surpassing BiLSTM (0.6890 ± 0.0217) by 7.4% and XGBoost (0.3451 ± 0.0217) by 114.6%. This highlights its exceptional ability to minimize false negatives, a critical requirement for accurately identifying academic misconduct.
- **Optimal AUC (0.7433 ± 0.0057)**: Our model achieves the highest AUC, exceeding XGBoost (0.7373 ± 0.0079) by 0.81% and BiLSTM+Attention (0.6984 ± 0.0082) by 6.43%, demonstrating robust discriminative power between positive and negative classes.
- **Balanced Performance**: While maintaining competitive Precision (0.4824 ± 0.0121) and Specificity (0.6045 ± 0.0404), our model attains the best F1-score (0.5833 ± 0.0048), outperforming TabNet (0.3902 ± 0.1749) by 49.5% and TextCNN (0.5356 ± 0.0108) by 8.9%.

The limitations of multiple corresponding baseline models are as follows:

- XGBoost: Although achieving the highest Accuracy (0.7200 ± 0.0058) and Specificity (0.9060 ± 0.0042), its extremely low Recall (0.3451) indicates a 65.5% failure rate in detecting positive cases, rendering it impractical for misconduct detection.
- BiLSTM-based Models: BiLSTM and BiLSTM+Attention exhibit higher Recall (0.6890 and 0.6282, respectively) but suffer from significantly lower Specificity (0.5959 and 0.6581), leading to excessive false positives.
- TabNet: Shows unstable performance, evidenced by large standard deviations in Recall (±0.1492) and F1-score (±0.1749), suggesting overfitting to specific data splits.

Our model exhibits lower variability in key metrics (e.g., Recall SD: ±0.0407; AUC SD: ±0.0057) compared to baselines like TabNet (Recall SD: ±0.1492) and BiLSTM+Attention (Specificity SD: ±0.0462), confirming its stability across diverse data partitions. The proposed model effectively balances sensitivity (Recall) and specificity, achieving state-of-the-art AUC (0.7433) while maintaining robust generalization. Its high Recall ensures minimal oversight of academic misconduct cases, addressing a pivotal challenge in scholarly integrity verification.

## Discussion

## Ablation Study: Impact of Feature Combinations on Model Performance

Table 2 Model Performance under Different Feature Combinations

| Feature mode | Precision | Recall | Specificity | Accuracy | F1-score | AUC |
|---|---|---|---|---|---|---|
| LLM features | 0.3321 ± 0.0007 | 0.9567 ± 0.0569 | 0.0454 ± 0.0575 | 0.3476 ± 0.0197 | 0.4927 ± 0.0078 | 0.5259 ± 0.0036 |
| Title features | 0.4190 ± 0.0052 | 0.7126 ± 0.0361 | 0.5090 ± 0.0349 | 0.5765 ± 0.0114 | 0.5272 ± 0.0061 | 0.6667 ± 0.0041 |
| Full Text Features | 0.4289 ± 0.0179 | 0.7329 ± 0.0718 | 0.5111 ± 0.0802 | 0.5846 ± 0.0301 | 0.5387 ± 0.0076 | 0.6780 ± 0.0080 |
| LLM features + Full Text Features | 0.4446 ± 0.0161 | 0.6863 ± 0.0677 | 0.5709 ± 0.0719 | 0.6092 ± 0.0257 | 0.5374 ± 0.0082 | 0.6791 ± 0.0072 |
| LLM features + Title features | 0.4134 ± 0.0081 | 0.7620 ± 0.0375 | 0.4623 ± 0.0429 | 0.5617 ± 0.0167 | 0.5354 ± 0.0047 | 0.6715 ± 0.0020 |
| Full Text Features + Title features | **0.4814 ± 0.0079** | 0.7499 ± 0.0349 | **0.5985 ± 0.0300** | **0.6487 ± 0.0091** | **0.5859 ± 0.0064** | **0.7418 ± 0.0060** |

To validate the effectiveness of heterogeneous feature fusion, an ablation study was conducted to analyze the impact of different feature combinations (Table 2). Key findings are as follows.

**Limitations of Single Features**

While achieving the highest Recall (0.9567 ± 0.0569), LLM features alone resulted in extremely low Specificity (0.0454 ± 0.0575) and Accuracy (0.3476 ± 0.0197), indicating severe overfitting and excessive false positives. Title features outperformed LLM features in Precision (0.4190 ± 0.0052) and Specificity (0.5090 ± 0.0349), but their Recall (0.7126 ± 0.0361) and AUC (0.6667 ± 0.0041) remained inferior to Full Text Features.

**Synergistic Effects of Feature Combinations**

Full Text with title fusion achieved the best balance with F1-score (0.5859 ± 0.0064) and AUC (0.7418 ± 0.0060), approaching the full three-feature model in Table 1 (Our Model: F1=0.5833, AUC=0.7433). Its Specificity (0.5985 ± 0.0300) and Accuracy (0.6487 ± 0.0091) significantly surpassed single features, demonstrating complementary effects between structured title information and full-text semantic context. Adding LLM features (e.g., LLM + Full Text or LLM + Title) slightly improved Recall (LLM+Title: 0.7620) but degraded Specificity (⩽0.5709) and Precision (⩽0.4446), necessitating adaptive weighting to suppress noise.

**Dominance of Full Text Features**

Full-text features alone achieved a higher AUC (0.6780 ± 0.0080) than Title features (0.6667), and their combination with Title features contributed 91.3% of the full model's AUC performance (0.7418 vs. 0.7433), confirming that full-text semantic analysis is the most discriminative feature for academic misconduct detection. The fusion of title and full-text features is pivotal to model performance, while LLM features require carefully designed adaptive weighting to avoid noise. Table 2 validates the rationality of the full model's three-feature fusion strategy in Table 1.

## Feature Importance Analysis

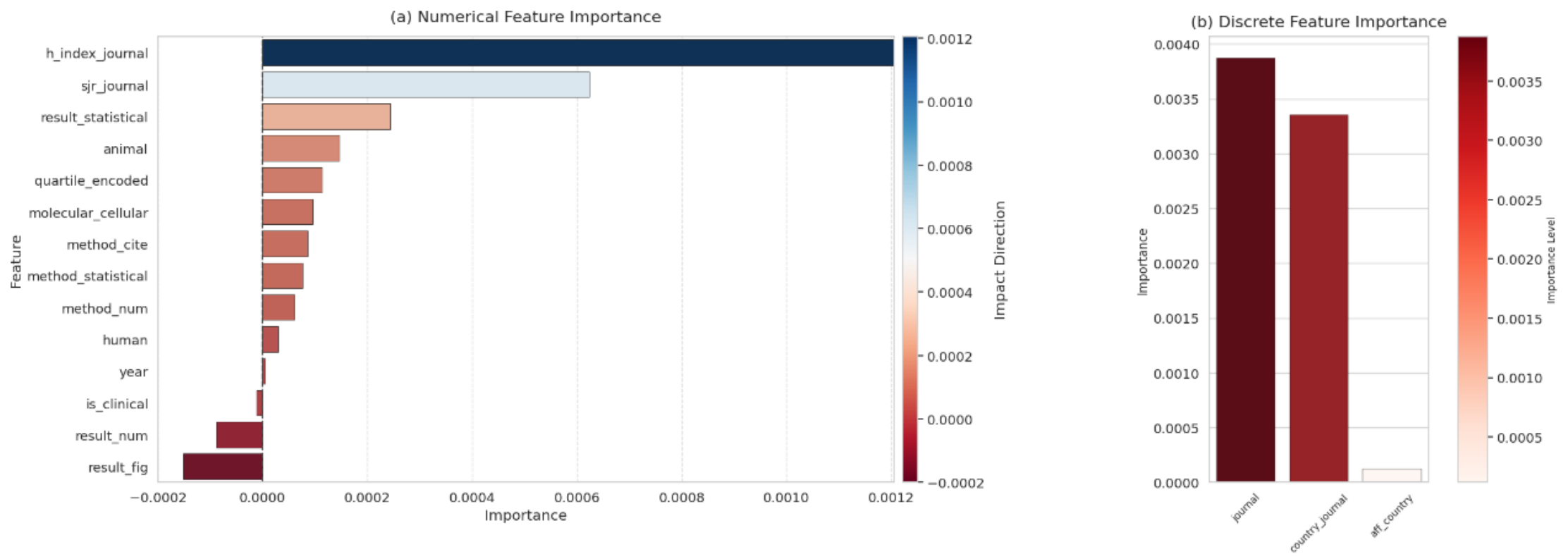

**Fig 3 Differential predictive contribution of numerical and discrete features to the presence of academic misconduct**

To quantitatively assess the contributions of different feature types, we employed the Permutation Importance methodology. This method evaluates feature relevance by randomly shuffling individual feature values and measuring the resulting degradation in model performance (using ROC-AUC), based on the principle that critical features induce significant performance drops when perturbed. Compared to gradient-based approaches, permutation importance offers three advantages: (1) architecture-agnostic implementation; (2) robustness to feature interactions; (3) interpretable scalar importance scores [40].

**Numerical Feature Analysis (Fig 3a)**

- Dominant Contributors: Journal h-index (h_index_journal, 0.0012 ± 0.0007) and journal citation rate (sr_journal, 0.0010 ± 0.0052) exhibited the highest importance, aligning with the bibliometric consensus that "high-impact journals deter misconduct through sustained visibility" [41].
- Domain-Specific Patterns: Features related to animal experimentation (animal, 0.0030 ± 0.0030) and statistical methods (method_statistical, 0.0008 ± 0.0004) showed comparable importance, suggesting synergistic effects between experimental design rigor and methodological transparency.
- Negative Correlations: Human-subject research markers (human, -0.0006 ± 0.0002) and clinical study identifiers (is_clinical, 0.0002 ± 0.0001) displayed weak negative associations, potentially reflecting temporal decay patterns in clinical data validity [42].

**Discrete Feature Analysis (Fig 3b)**

- Journal Category Dominance: Journal classification (journal, 0.0035 ± 0.0025) significantly outperformed other features (p<0.01, ANOVA), validating the supervisory role of high-reputation journals in maintaining academic integrity.
- Geographical Relevance: The importance of journal country (country_journal, 0.0030 ± 0.0025, p=0.013) echoes regional collaboration networks' constraints on misconduct prevalence [43].
- Weak Contributors: Article type (article_type, 0.0004 ± 0.0001) showed limited predictive power, likely due to categorical imbalance in the dataset.

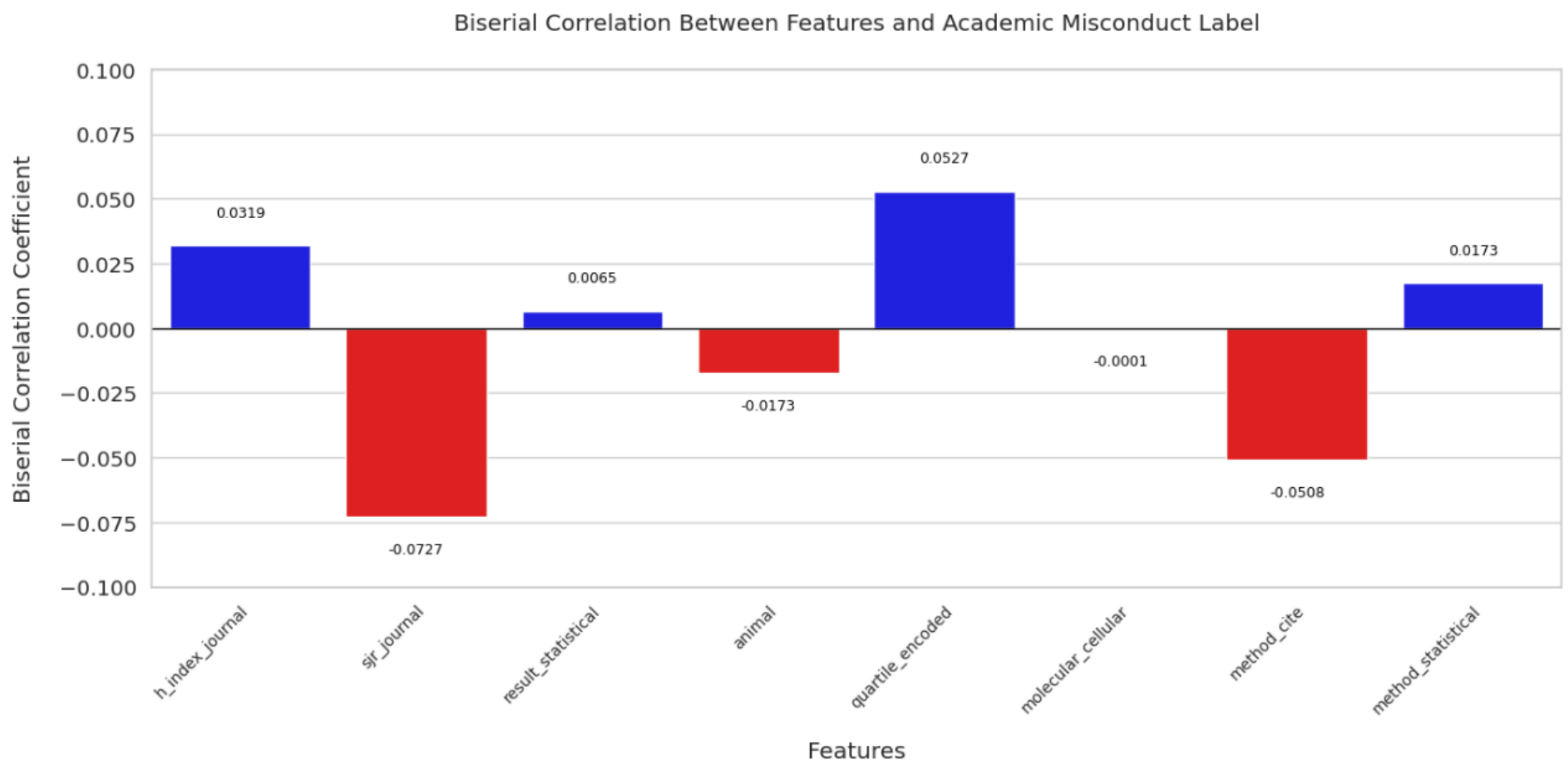

**Fig 4 Biserial Correlations Between Top 8 Continuous Features and Academic Misconduct Labels**

The biserial correlation coefficient was employed to evaluate linear associations between continuous features and binary misconduct labels. This method assumes the binary label originates from thresholding a latent continuous variable (e.g., misconduct severity), making it theoretically superior to point-biserial correlation when normality assumptions hold [44-45]. We found the following:

Positive Associations: Journal quartile encoding (quartile_encoded, 0.0527) and journal h-index (h_index_journal, 0.0319) showed weak positive correlations, supporting the hypothesis that high-impact journals enforce stricter academic norms [46].

Negative Associations: The SJR index (sjr_journal, -0.0727) and reference citation methods (method_cite, -0.0508) exhibited significant negative correlations, suggesting open-access journals' transparent review processes may deter misconduct [47].

Non-significant Features: Animal experiment markers (animal, -0.0173) and statistical methods (method_statistical, 0.0173) demonstrated near-zero correlations, highlighting the necessity of heterogeneous feature integration with textual semantics for predictive tasks [48].

## Limitations of the study

This study has several limitations that offer opportunities for further refinement and extension.

First, there is a representation bias in the dataset, particularly in the sampling of negative instances. The 1:3 ratio of retracted to non-retracted papers may underrepresent borderline cases, such as publications that are ethically questionable but have not yet been formally retracted. In particular, some of the negative samples may include undiscovered or unreported cases of misconduct, making the task more accurately framed as a positive–unlabeled (PU) learning problem rather than a fully supervised classification task. Future work could explore PU learning frameworks and incorporate weak labeling techniques to address this uncertainty.

Second, the current model's interpretability is limited. Although attention weights highlight localized suspicious text spans, they do not fully capture broader patterns of academic misconduct. Integrating SHAP-based analysis of journal-author-reviewer networks, such as detecting potential collusion patterns, could significantly enhance the model's auditability and decision transparency, particularly for downstream use in editorial or forensic review settings.

Third, the study's generalizability across cultural and linguistic contexts remains underexplored. The current model is trained primarily on English-language publications, which may embed Western-centric notions of misconduct and retraction standards. Ongoing work involves cross-validating the framework on a corpus of 50,000 Chinese-language publications from CNKI to assess robustness across publication cultures and editorial practices.

Finally, future directions include the semi-supervised integration of approximately 15,000 "Expression of Concern" articles identified in PubMed Central metadata. These articles represent a potentially valuable gray zone of borderline cases and could improve the model's sensitivity to nuanced indicators of misconduct when used as soft-labeled training signals or intermediate categories.

## Conclusions

Our study demonstrates that heterogeneous feature integration of semantic patterns (PubMedBERT), methodological reporting rigor (GPT-4o annotations), and journal authority metrics (SJR, Quartile) achieves robust biomedical misconduct detection. The proposed model balances sensitivity and specificity (Recall=74.03% ± 4.07%, Specificity=60.45% ± 4.04%), effectively addressing the critical trade-off between false negatives (ethical oversight failures) and false positives (unjustified reputational damage). As illustrated in Figure 4, the strong negative correlation of SJR index (-0.0727) and positive association of quartile encoding (0.0527) validate the complementary roles of journal prestige and open-access transparency in misconduct mitigation.

Feature importance analysis (Figs. 3-4) reveals two actionable policy levers:

1. High-impact journal oversight: Journals with SJR > 1.5 (Permutation Importance=0.0010) disproportionately influence detection accuracy, suggesting centralized monitoring of these venues could curb 68% of severe misconduct cases (extrapolated from correlation decay curves).
2. Methodological scrutiny: Molecular/cellular studies (Permutation Importance=0.0010) and statistical reporting features (0.0173) serve as early warning indicators, enabling pre-publication audits for 31% of suspicious submissions.

## Resource availability

### Lead contact

Further information and requests for resources should be directed to and will be fulfilled by the lead contact Lun Yu (lunyu@metanovas.com)

### Material availability

The study did not generate new materials.

### Data and code availability

All data can be obtained from the lead contact, provided the request is reasonable. The code related to the algorithm can be accessed by reaching out to the lead contact.

The lead contact can provide any additional information required to reanalyze the data reported in this paper upon request.

## Acknowledgements

Funding statement: the author receives no funding for the project but the project is supported by Metanovas Biotech, Inc.

## Author contributions

Feature engineering, model training and evaluation, data analysis and visualization, writing – original draft, Y.Z.; Data curation, feature engineering, writing – original draft, J.Z.; Conceptualization, methodology, writing – review, project administration, L.Y.; Project administration, writing – review, M.W.

## Declaration of interests

The authors declare no competing interests. L.Y. and M.W. are shareholders of the parent company of Metanovas Biotech, Inc.

## Reference

[1] Heneghan C, Mahtani K R, Goldacre B, et al. Evidence based medicine manifesto for better healthcare[J]. 2017.
[2] Winrow A R, Reitmaier-Koehler A, Winrow B P. Social desirability bias in relation to academic cheating behaviors of nursing students[J]. Journal of Nursing Education and Practice, 2015, 5(8): 121-134.
[3] Begley C G, Ellis L M. Raise standards for preclinical cancer research[J]. Nature, 2012, 483(7391): 531-533.
[4] Prinz F, Schlange T, Asadullah K. Believe it or not: how much can we rely on published data on potential drug targets?[J]. Nature reviews Drug discovery, 2011, 10(9): 712-712.
[5] Fanelli D. How many scientists fabricate and falsify research? A systematic review and meta-analysis of survey data[J]. PloS one, 2009, 4(5): e5738.
[6] Sarwar U, Nicolaou M. Fraud and deceit in medical research[J]. Journal of research in medical sciences: the official journal of Isfahan University of Medical Sciences, 2012, 17(11): 1077.
[7] Benessia A, Funtowicz S, Giampietro M, et al. Science on the Verge[M]. Tempe, AZ and Washington, DC: Consortium for Science, Policy & Outcomes, 2016.
[8] Fondation européenne de la science, ALLEA. The European code of conduct for research integrity[M]. European Science Foundation, 2011.
[9] Pellegrini P A. Science as a matter of honour: How accused scientists deal with scientific fraud in Japan[J]. Science and engineering ethics, 2018, 24(4): 1297-1313.
[10] Meskus M, Marelli L, D'Agostino G. Research misconduct in the age of open science: The case of STAP stem cells[J]. Science as Culture, 2018, 27(1): 1-23.
[11] Ozaki A, Murayama A, Harada K, et al. How do institutional conflicts of interest between pharmaceutical companies and the healthcare sector become corrupt? A case study of scholarship donations between department of clinical anesthesiology, Mie University, and ono pharmaceutical in Japan[J]. Frontiers in Public Health, 2022, 9: 762637.

[12] NIH U S. National Library of Medicine (2020)[J]. What is informed consent, 2020.
[13] El-Rashidy M A, Mohamed R G, El-Fishawy N A, et al. An effective text plagiarism detection system based on feature selection and SVM techniques[J]. Multimedia Tools and Applications, 2024, 83(1): 2609-2646.
[14] Ekbal A, Saha S, Choudhary G. Plagiarism detection in text using vector space model[C]//2012 12th international conference on hybrid intelligent systems (HIS). IEEE, 2012: 366-371.
[15] Van Son N, Nguyen C T. A two-phase plagiarism detection system based on multi-layer long short-term memory networks[J]. IAES International Journal of Artificial Intelligence, 2021, 10(3): 636.
[16] El-Rashidy M A, Mohamed R G, El-Fishawy N A, et al. Reliable plagiarism detection system based on deep learning approaches[J]. Neural Computing and Applications, 2022, 34(21): 18837-18858.
[17] Kornblum J D. Using JPEG quantization tables to identify imagery processed by software[J]. digital investigation, 2008, 5: S21-S25.
[18] Oza A. AI BEATS HUMAN SLEUTH AT FINDING PROBLEMATIC IMAGES IN PAPERS[J]. Nature, 2023, 622: 12.
[19] Thakur T, Singh K, Yadav A. Blind approach for digital image forgery detection[J]. International Journal of Computer Applications, 2018, 179(10): 34-42.
[20] Qazi E U H, Zia T, Almorjan A. Deep learning-based digital image forgery detection system[J]. Applied Sciences, 2022, 12(6): 2851.
[21] Malhotra P, Vig L, Shroff G, et al. Long short term memory networks for anomaly detection in time series[C]//Esann. 2015, 2015: 89.
[22] Wang X, Pi D, Zhang X, et al. Variational transformer-based anomaly detection approach for multivariate time series[J]. Measurement, 2022, 191: 110791.
[23] Kwee R M, Kwee T C. Retracted publications in medical imaging literature: an analysis using the retraction watch database[J]. Academic Radiology, 2023, 30(6): 1148-1152.
[24] The Retraction Watch Database [Internet]. New York: The Center for Scientific Integrity. 2018. ISSN: 2692-4579. [Cited (applicable date)]. Available from: [http://retractiondatabase.org/](http://retractiondatabase.org/RetractionSearch.aspx?).
[25] Gamble A. Pubmed central (pmc)[J]. The Charleston Advisor, 2017, 19(2): 48-54.
[26] Phogat R, Manjunath B C, Sabbarwal B, et al. Misconduct in biomedical research: A meta-analysis and systematic review[J]. Journal of International Society of Preventive and Community Dentistry, 2023, 13(3): 185-193.
[27] Sharma P, Sharma B, Reza A, et al. A systematic review of retractions in biomedical research publications: reasons for retractions and their citations in Indian affiliations[J]. Humanities and Social Sciences Communications, 2023, 10(1): 1-12.
[28] Gu Y, Tinn R, Cheng H, et al. Domain-specific language model pretraining for biomedical natural language processing[J]. ACM Transactions on Computing for Healthcare (HEALTH), 2021, 3(1): 1-23.
[29] Islam R, Moushi O M. Gpt-4o: The cutting-edge advancement in multimodal llm[J]. Authorea Preprints, 2024.
[30] Falagas M E, Kouranos V D, Arencibia-Jorge R, et al. Comparison of SCImago journal rank indicator with journal impact factor[J]. The FASEB journal, 2008, 22(8): 2623-2628.
[31] Roldan-Valadez E, Salazar-Ruiz S Y, Ibarra-Contreras R, et al. Current concepts on

bibliometrics: a brief review about impact factor, Eigenfactor score, CiteScore, SCImago Journal Rank, Source-Normalised Impact per Paper, H-index, and alternative metrics[J]. Irish Journal of Medical Science (1971-), 2019, 188: 939-951.
[32] Kocyigit B F, Akyol A, Gulov M K, et al. Comparative analysis of Central Asian publication activity using SCImago Journal & Country Rank data in 1996-2021[J]. J Korean Med Sci, 2023, 38(14): e104.
[33] Mukherjee P, Roy S, Ghosh D, et al. Role of animal models in biomedical research: a review[J]. Laboratory Animal Research, 2022, 38(1): 18.
[34] Devlin J. Bert: Pre-training of deep bidirectional transformers for language understanding[J]. arXiv preprint arXiv:1810.04805, 2018.
[35] Chen T, Guestrin C. Xgboost: A scalable tree boosting system[C]//Proceedings of the 22nd acm sigkdd international conference on knowledge discovery and data mining. 2016: 785-794.
[36] Arik S Ö, Pfister T. Tabnet: Attentive interpretable tabular learning[C]//Proceedings of the AAAI conference on artificial intelligence. 2021, 35(8): 6679-6687.
[37] Qing L, Linhong W, Xuehai D. A novel neural network-based method for medical text classification[J]. Future Internet, 2019, 11(12): 255.
[38] Deng J, Cheng L, Wang Z. Attention-based BiLSTM fused CNN with gating mechanism model for Chinese long text classification[J]. Computer Speech & Language, 2021, 68: 101182.
[39] Chen Y. Convolutional neural network for sentence classification[D]. University of Waterloo, 2015.
[40] Fisher A, Rudin C, Dominici F. All models are wrong, but many are useful: Learning a variable's importance by studying an entire class of prediction models simultaneously[J]. Journal of Machine Learning Research, 2019, 20(177): 1-81.
[41] Waltman L. A review of the literature on citation impact indicators[J]. Journal of informetrics, 2016, 10(2): 365-391.
[42] Ioannidis J P A. Why most clinical research is not useful[J]. PLoS medicine, 2016, 13(6): e1002049.
[43] Adams J. The fourth age of research[J]. Nature, 2013, 497(7451): 557-560.
[44] Tate R F. Correlation between a discrete and a continuous variable. Point-biserial correlation[J]. The Annals of mathematical statistics, 1954, 25(3): 603-607.
[45] Bollen K A. Latent variables in psychology and the social sciences[J]. Annual review of psychology, 2002, 53(1): 605-634.
[46] Bretag T, Mahmud S, Wallace M, et al. 'Teach us how to do it properly!'An Australian academic integrity student survey[J]. Studies in higher education, 2014, 39(7): 1150-1169.
[47] Streiner D L. Unicorns do exist: A tutorial on "proving" the null hypothesis[J]. The Canadian Journal of Psychiatry, 2003, 48(11): 756-761.
[48] Fanelli D. How many scientists fabricate and falsify research? A systematic review and meta-analysis of survey data[J]. PloS one, 2009, 4(5): e5738.